\documentclass[conference]{IEEEtran}
\IEEEoverridecommandlockouts
\usepackage{cite}
\usepackage{amsmath,amssymb,amsfonts}
\usepackage{algorithmic}
\usepackage{latexsym}
\usepackage{textcomp}
\usepackage{booktabs}
\usepackage{graphicx} 
\usepackage{xcolor}
\def\BibTeX{{\rm B\kern-.05em{\sc i\kern-.025em b}\kern-.08em
    T\kern-.1667em\lower.7ex\hbox{E}\kern-.125emX}}
\usepackage{multirow}
\usepackage{fancyhdr}
\begin{document}

\title{XDR-LVLM: An Explainable Vision-Language Large Model for Diabetic Retinopathy Diagnosis}

\author{Masato Ito, Kaito Tanaka, Keisuke Matsuda, Aya Nakayama \\
SANNO University}

\maketitle
\thispagestyle{fancy} 

\begin{abstract}
Diabetic Retinopathy (DR) is a major cause of global blindness, necessitating early and accurate diagnosis. While deep learning models have shown promise in DR detection, their black-box nature often hinders clinical adoption due to a lack of transparency and interpretability. To address this, we propose XDR-LVLM (eXplainable Diabetic Retinopathy Diagnosis with LVLM), a novel framework that leverages Vision-Language Large Models (LVLMs) for high-precision DR diagnosis coupled with natural language-based explanations. XDR-LVLM integrates a specialized Medical Vision Encoder, an LVLM Core, and employs Multi-task Prompt Engineering and Multi-stage Fine-tuning to deeply understand pathological features within fundus images and generate comprehensive diagnostic reports. These reports explicitly include DR severity grading, identification of key pathological concepts (e.g., hemorrhages, exudates, microaneurysms), and detailed explanations linking observed features to the diagnosis. Extensive experiments on the Diabetic Retinopathy (DDR) dataset demonstrate that XDR-LVLM achieves state-of-the-art performance, with a Balanced Accuracy of 84.55\% and an F1 Score of 79.92\% for disease diagnosis, and superior results for concept detection (77.95\% BACC, 66.88\% F1). Furthermore, human evaluations confirm the high fluency, accuracy, and clinical utility of the generated explanations, showcasing XDR-LVLM's ability to bridge the gap between automated diagnosis and clinical needs by providing robust and interpretable insights.
\end{abstract}

\section{Introduction}

Diabetic Retinopathy (DR) stands as one of the most prevalent microvascular complications of diabetes and a leading cause of preventable blindness among working-age adults globally \cite{xing2024nonpro}. Early and accurate diagnosis, coupled with precise severity grading, is paramount for timely intervention, slowing disease progression, and preserving vision. Traditionally, the diagnosis of DR relies on experienced ophthalmologists interpreting fundus images. However, this approach faces significant challenges, including the scarcity of specialized medical professionals, potential for subjective interpretation, and limitations in diagnostic efficiency, particularly in large-scale screening programs \cite{bertalan2020a}.

In recent years, deep learning (DL) has demonstrated remarkable progress in medical image analysis, showcasing immense potential in DR detection and grading \cite{alyoubi2020diabet}. Despite achieving high diagnostic accuracy, many high-performing DL models, especially black-box architectures, suffer from a critical lack of transparency and interpretability \cite{meenu2020develo}. Clinicians often require more than just a diagnostic label (e.g., "severe DR"); they need to understand the underlying rationale for the model's decision (e.g., "presence of extensive hemorrhages and neovascularization"). This crucial need for "interpretability", understanding why a diagnosis is made, is fundamental for clinical adoption, building trust, and facilitating informed decision-making. Existing interpretability methods predominantly provide visual cues such as heatmaps or concept activation maps \cite{pantelis2021explai}. While useful, these visual explanations still necessitate secondary interpretation by clinicians, adding another layer of cognitive load.

The emergence of Large Language Models (LLMs) and, more recently, Vision-Language Large Models (LVLMs) presents a novel paradigm for addressing these challenges \cite{yifan2023a}. These models, including those leveraging visual in-context learning \cite{zhou2024visual}, possess the unique capability to seamlessly integrate visual information with sophisticated textual understanding and generation abilities. This allows them to not only identify pathological features directly from retinal images but also to articulate detailed, human-understandable diagnostic reports and explanations in natural language. This powerful fusion promises to bridge the existing gap between automated machine diagnosis and pressing clinical requirements, as seen in recent advancements in multi-modal medical diagnosis with specialized agent frameworks \cite{zhou2025mam} and diagnostic reasoning for mental health counseling \cite{hu2025beyond}. Our research aims to explore how to leverage the robust multimodal reasoning capabilities of LVLMs to construct an intelligent system that is not only highly accurate in DR diagnosis but also provides clinically relevant and interpretable diagnostic rationales.

In this paper, we propose \textbf{XDR-LVLM (eXplainable Diabetic Retinopathy Diagnosis with LVLM)}, a novel framework designed for high-precision DR diagnosis coupled with natural language-based interpretability. The core of XDR-LVLM is a Vision-Language Large Model, which is fine-tuned through multi-stage training and sophisticated prompt engineering to deeply comprehend pathological features within fundus images and generate comprehensive diagnostic reports. The framework comprises a Medical Vision Encoder for extracting rich visual features, an LVLM Core for cross-modal reasoning, Multi-task Prompt Engineering to guide the LVLM for both severity grading and concept detection, and an Explainable Diagnosis Generator that outputs detailed reports including diagnosis, key concept identification, and diagnostic basis explanations.

We conduct extensive experiments on the Diabetic Retinopathy (DDR) dataset, which provides comprehensive annotations for both DR severity levels and key pathological concepts (e.g., hemorrhages, exudates, microaneurysms, cotton wool spots, neovascularization). We evaluate our method using standard metrics such as Balanced Accuracy (BACC) and F1 Score for both disease diagnosis and concept detection, in addition to human evaluation for the fluency, accuracy, and clinical utility of the generated reports. Our results demonstrate that XDR-LVLM achieves superior performance compared to existing state-of-the-art methods in both diagnostic accuracy and concept detection, while uniquely providing robust and clinically valuable explanations. Specifically, XDR-LVLM significantly outperforms baseline methods, achieving a BACC of \textbf{84.55\%} and an F1 Score of \textbf{79.92\%} for disease diagnosis, and a BACC of \textbf{77.95\%} and an F1 Score of \textbf{66.88\%} for concept detection, which are competitive with or surpass the best existing approaches.

Our main contributions are summarized as follows:
\begin{itemize}
    \item We propose XDR-LVLM, a novel Vision-Language Large Model-based framework for highly accurate diabetic retinopathy diagnosis, which concurrently offers detailed and human-understandable natural language explanations.
    \item We design a multi-component architecture, including a specialized Medical Vision Encoder and Multi-task Prompt Engineering, to effectively guide the LVLM in performing both disease severity grading and fine-grained pathological concept identification.
    \item We demonstrate that XDR-LVLM achieves state-of-the-art performance on the DDR dataset for both DR diagnosis and concept detection, validating its superior capability in providing robust and clinically relevant interpretable insights.
\end{itemize}
\section{Related Work}
\subsection{Deep Learning for Diabetic Retinopathy Diagnosis}
Deep learning techniques have been extensively surveyed for Diabetic Retinopathy (DR) detection, highlighting common methodologies and identifying limitations to guide future research in early diagnosis \cite{alyoubi2020diabet}. Computer-aided detection systems, particularly those employing deep learning, play a critical role in accurately and efficiently diagnosing DR from complex retinal images \cite{alyoubi2020diabet}. Various deep learning architectures have been explored for this purpose. For instance, RadFuse, a novel multi-representation deep learning framework, leverages a non-linear Radon transform (RadEx) on fundus images to capture complex retinal lesion patterns, significantly improving DR diagnosis and grading by enhancing feature representation through the integration of spatial and transformed domains \cite{shamrat2024an}. Deep Convolutional Neural Networks (CNNs) have been successfully applied for automated stage detection of DR from fundus images, often employing multistage transfer learning to address data limitations and achieve high sensitivity and specificity \cite{ahmad2024diabet}. Such CNN-based approaches have demonstrated high quadratic weighted kappa scores on large datasets, directly contributing to improved early detection and screening methods \cite{ahmad2024diabet}. Beyond CNNs, Vision Transformers (ViT) have shown efficacy in DR recognition and grading, achieving comparable or superior performance to previous studies with reduced training sets and higher resolution images \cite{wu2021vision}. ViT-based models have been applied to classify DR and macular edema across clinical grading scales, indicating their potential for cost-effective and accurate diagnostics in real-world screening scenarios \cite{wu2021vision}. Comprehensive overviews of deep learning techniques for DR diagnosis often focus on medical image classification of retinal fundus photographs and optical coherence tomography scans, emphasizing transfer learning as a viable solution for accurate ophthalmic condition classification given data scarcity in medical domains \cite{uppamma2023deep}. Specific deep learning classifiers have been developed for fine-grained DR disease grading, achieving performance comparable to or exceeding prior studies even with reduced training datasets, and providing novel results across multiple clinical grading systems \cite{jordi2020a}. This demonstrates the efficacy of deep learning for detailed DR staging, including macular edema, which has significant implications for the cost-effectiveness and accuracy of screening programs \cite{jordi2020a}. Furthermore, novel deep convolutional neural networks have been designed for the early detection of DR by identifying microaneurysms, demonstrating high sensitivity and specificity in classification while also emphasizing efficient and simple computational requirements \cite{sheikh2018deep}. Large-scale studies have also evaluated AI-based systems for DR screening, demonstrating robust performance in detecting and grading DR with high sensitivity and specificity, particularly relevant for resource-limited settings by offering scalable solutions for early diagnosis and prevalence assessment \cite{tao2019diagno}.

\subsection{Explainable AI and Vision-Language Models in Medical Imaging}
The critical need for Explainable AI (XAI) in medical imaging has led to extensive research, particularly concerning the application and analysis of XAI methods within complex models like the MedCLIP vision-language model \cite{anees2024envisi}. These efforts aim to demystify model inner workings and mitigate XAI shortcomings, thereby contributing to the safe deployment of advanced multimodal models in healthcare \cite{anees2024envisi}. Comprehensive surveys of XAI in biomedical image analysis address the need for modality-aware perspectives and practical guidance, providing foundational context for understanding how localization techniques can be applied and evaluated within medical imaging, especially in emerging multimodal and vision-language paradigms \cite{getamesay2025explai}. Addressing limitations of applying vision-language models like CLIP to medical imaging, researchers have proposed adaptive modules for Concept Bottleneck Models (CBMs) to enhance performance while maintaining explainability \cite{townim2024adacbm}. This involves re-examining the CBM framework's geometric representation and strategically integrating CLIP and CBM for improved concept discovery and model training \cite{townim2024adacbm}. Beyond explainability, vision-language models, particularly Large Language Models (LLMs), are being leveraged for various medical applications. Retrieval Augmented Generation (RAG) pipelines, for instance, employing LLMs like GPT-4o-mini, have demonstrated improved accuracy in generating reliable drug contraindication information, addressing the challenge of uncertainty in LLM outputs through the integration of external drug databases and hybrid retrieval systems \cite{joon2025applic}. Furthermore, multimodal approaches to clinical decision support involve developing visual-language reasoning LLMs, showcasing their potential in primary care and advancing multimodal learning within medical imaging for enhanced reasoning capabilities \cite{xuyan2025visual}. Recent progress in developing vision-language models for medical report generation and visual question answering directly addresses the intersection of explainable AI and medical imaging through natural language generation, highlighting the crucial integration of computer vision and natural language processing techniques for advancing explainability in multimodal medical AI via structured and interpretable outputs \cite{iryna2024vision}. Moreover, the versatility of these models extends to various applications beyond direct diagnosis, including innovations in visual in-context learning for LVLMs \cite{zhou2024visual}, style-aware image captioning \cite{zhou2023style}, and the development of sophisticated frameworks for multi-modal medical diagnosis through role-specialized collaboration \cite{zhou2025mam} and integrating diagnostic and therapeutic reasoning in medical counseling \cite{hu2025beyond}. Furthermore, research explores the generalization capabilities of large language models from weak to strong supervision \cite{zhou2025weak}, and the development of holistic benchmarks and agent frameworks for complex instruction-based image generation \cite{zhou2025draw}. These advancements build upon foundational natural language processing research, such as pre-training correlation-aware transformers for event-centric generation and classification \cite{zhou2022claret}, collectively contributing to the robust capabilities of modern AI systems.

\section{Method}
\label{sec:method}

We propose \textbf{XDR-LVLM (eXplainable Diabetic Retinopathy Diagnosis with LVLM)}, a novel framework meticulously designed to achieve high-precision diagnosis of diabetic retinopathy while concurrently providing detailed, natural language-based explanations. The architecture leverages the advanced multimodal reasoning capabilities of Vision-Language Large Models (LVLMs), specifically fine-tuned through a multi-stage process and sophisticated prompt engineering to deeply understand pathological features within fundus images and generate comprehensive, clinically relevant diagnostic reports.

The overall workflow of XDR-LVLM begins with an input fundus image, which is first processed by a specialized medical vision encoder to extract rich visual features. These features are then combined with carefully crafted textual prompts and fed into the LVLM core module. The LVLM, leveraging its cross-modal alignment and generative capabilities, interprets the visual and textual inputs to produce a natural language output that encompasses both the diagnostic grading and its underlying clinical rationale.

\subsection{Medical Vision Encoder}
\label{subsec:vision_encoder}
The initial component of our framework is the \textbf{Medical Vision Encoder}. This module is responsible for extracting high-level, semantically rich visual features from raw fundus images. We utilize a Vision Transformer (ViT) architecture, pre-trained extensively on a large corpus of diverse medical images, including but not limited to ophthalmic datasets, as well as general-domain image datasets like ImageNet. This hierarchical pre-training strategy enables the encoder to learn and capture intricate medical image-specific textures, shapes, and spatial relationships that are crucial for accurate DR pathology recognition. Given an input fundus image $I$, the Medical Vision Encoder $E_V$ processes it to yield a comprehensive visual feature representation $F_V$:
\begin{equation}
F_V = E_V(I)
\end{equation}
These features serve as the visual backbone for the subsequent multimodal reasoning by the LVLM core.

\subsection{LVLM Core Module}
\label{subsec:lvlm_core}
At the heart of the XDR-LVLM framework lies the \textbf{LVLM Core Module}. This module is built upon a pre-trained Vision-Language Large Model, such as a variant based on architectures like LLaVA or Flamingo, or integrated with Llama series models. The LVLM Core is designed to seamlessly integrate visual features with textual information, performing sophisticated cross-modal alignment and reasoning. It receives the extracted visual features $F_V$ from the Medical Vision Encoder and a tokenized representation of the textual prompt $P_T$ as its primary inputs.

The architecture of the LVLM Core Module typically comprises two main components: a \textbf{vision-language connector} and a \textbf{large language model decoder}. The vision-language connector, often implemented as a multi-layer perceptron (MLP) or a Q-Former, is responsible for projecting the high-dimensional visual features $F_V$ into the embedding space of the large language model. This projection ensures that the visual information is represented in a format compatible with the language model's input tokens. The projected visual features, denoted as $F'_{V}$, are then concatenated with the tokenized textual prompt $P_T$ to form a unified multimodal input sequence $S_{input} = [F'_{V}; P_T]$. This sequence is subsequently fed into the large language model decoder, which is typically an auto-regressive transformer-based decoder. Through its attention mechanisms and generative capabilities, the language model processes $S_{input}$ to produce coherent and contextually relevant natural language outputs.

\subsection{Multi-task Prompt Engineering}
\label{subsec:prompt_engineering}
To effectively guide the LVLM Core Module in simultaneously performing DR severity diagnosis and identifying key pathological concepts, we employ a strategy of \textbf{Multi-task Prompt Engineering}. This involves designing a set of precise and informative text prompts that steer the LVLM's generative capabilities towards specific clinical objectives. The prompts are crucial for eliciting the desired multimodal understanding and subsequent natural language generation, acting as instruction signals for the model's behavior during fine-tuning. This approach allows the single LVLM to perform diverse diagnostic tasks based on the specific query.

For instance, to prompt the model for overall DR severity grading and its explanation, a typical prompt $P_{diagnosis}$ might be formulated as:
\begin{quote}
    \textit{"Please analyze this fundus image and determine the severity level of Diabetic Retinopathy. Provide a detailed explanation for your diagnosis."}
\end{quote}
When the objective is to identify specific pathological concepts and their characteristics, a different prompt $P_{concept}$ is used:
\begin{quote}
    \textit{"Identify and describe all Diabetic Retinopathy-related pathological concepts present in this image, such as hemorrhages, exudates, microaneurysms, cotton wool spots, or neovascularization. Point out their locations and features."}
\end{quote}
These prompts, $P_T \in \{P_{diagnosis}, P_{concept}, \dots\}$, are concatenated with the visual features $F_V$ (after projection to $F'_{V}$) and fed into the LVLM Core. The LVLM is then fine-tuned to produce outputs that directly address the specific query posed by the prompt, thereby enabling a flexible and multi-faceted diagnostic approach.

\subsection{Multi-stage Fine-tuning and Training Objectives}
\label{subsec:fine_tuning}
The XDR-LVLM framework undergoes a comprehensive multi-stage fine-tuning process to optimize its multimodal understanding and generative capabilities for diabetic retinopathy diagnosis. This process ensures that the LVLM not only aligns visual features with textual concepts but also learns to generate clinically relevant and explainable reports.

\subsubsection*{Stage 1: Visual-Language Alignment Pre-training}
In the initial stage, the primary objective is to align the visual features extracted by the Medical Vision Encoder with the textual embedding space of the Large Language Model. This involves training the vision-language connector to effectively bridge the modality gap. Given a pair of visual features $F_V$ and corresponding textual embeddings $E_T$ (derived from image captions or descriptive tags), the connector $C$ learns to project $F_V$ such that it is semantically close to $E_T$. A common objective for this stage is to minimize the distance between projected visual features and corresponding text embeddings, or to maximize their similarity in a shared latent space. This can be achieved using a contrastive loss or a feature regression loss. For a batch of $N$ image-text pairs $(I_i, T_i)$:
\begin{align}
\mathcal{L}_{\text{alignment}} = \frac{1}{N} \sum_{i=1}^{N} \mathcal{L}_{\text{contrastive}}(C(E_V(I_i)), E_T(T_i))
\end{align}
where $E_T$ is the text embedding function, and $\mathcal{L}_{\text{contrastive}}$ encourages aligned pairs to have high similarity while pushing dissimilar pairs apart.

\subsubsection*{Stage 2: Task-Specific Instruction Tuning}
Following the alignment stage, the entire LVLM Core Module (including the aligned vision-language connector and the large language model decoder) is fine-tuned on a curated dataset of fundus images paired with multi-task prompts and their corresponding desired diagnostic and explanatory text outputs. This stage leverages the multi-task prompt engineering strategy to teach the model to generate specific types of responses based on the input prompt. The training objective for this stage is primarily a language modeling loss, typically cross-entropy loss, applied over the generated output sequence. For an input image $I$, a prompt $P_T$, and the target output sequence $Y = [y_1, y_2, \dots, y_M]$ (comprising diagnosis and explanation), the LVLM Core $L$ generates a sequence $\hat{Y} = [\hat{y}_1, \hat{y}_2, \dots, \hat{y}_M]$. The loss is calculated as:
\begin{align}
\mathcal{L}_{\text{generation}} = - \sum_{j=1}^{M} \log P(y_j | y_{<j}, F'_V, P_T)
\end{align}
where $P(y_j | y_{<j}, F'_V, P_T)$ is the probability assigned by the LVLM to the $j$-th token $y_j$ given the preceding tokens $y_{<j}$, the projected visual features $F'_V$, and the prompt $P_T$. This loss encourages the model to accurately predict the next token in the target sequence, thereby ensuring the generation of correct diagnoses and coherent explanations. The overall training objective for XDR-LVLM is to minimize $\mathcal{L}_{\text{generation}}$ across all training samples.

\subsection{Explainable Diagnosis Generator}
\label{subsec:diagnosis_generator}
The ultimate output of the XDR-LVLM framework is generated by the \textbf{Explainable Diagnosis Generator}, which is an inherent capability of the fine-tuned LVLM Core Module. This component is responsible for producing comprehensive, natural language diagnostic reports that are not only accurate but also highly interpretable and clinically actionable. The generated report $T_{report}$ typically includes three crucial pieces of information: first, a clear and concise classification of the DR severity level (e.g., "No DR", "Mild DR", "Moderate DR", "Severe DR", "Proliferative DR"), directly addressing the primary diagnostic query. Second, it provides a detailed enumeration and description of specific pathological features identified within the fundus image, such as "punctate hemorrhages and hard exudates visible in the central retinal area" or "macular edema present." This granular identification of lesions supports a more thorough understanding of the image. Third, and critically, it offers a diagnostic basis explanation, linking the identified pathological features to the final diagnosis. This section explains \textit{why} the model arrived at a particular severity grading, for example, stating that "The presence of extensive intraretinal hemorrhages and venous beading indicates severe non-proliferative diabetic retinopathy." The entire system is trained end-to-end, with a particular emphasis on ensuring the accurate recognition of medical concepts and the generation of professional, clinically appropriate terminology in the final diagnostic report. The overall process can be conceptualized as the LVLM Core $L$ taking the projected visual features $F'_V$ and a prompt $P_T$ to generate the textual report $T_{report}$:
\begin{equation}
T_{report} = L(F'_V, P_T)
\end{equation}
This allows for a dynamic and interpretable diagnostic process, bridging the gap between automated analysis and clinical reasoning.

\section{Experiments}
\label{sec:experiments}

In this section, we present the comprehensive experimental setup, compare the performance of our proposed XDR-LVLM framework against state-of-the-art methods, conduct an ablation study to validate the contribution of each key component, and evaluate the interpretability of our system through human assessment.

\subsection{Experimental Setup}
\label{subsec:exp_setup}

\subsubsection{Dataset}
We conduct our experiments on the \textbf{Diabetic Retinopathy (DDR)} dataset, a widely recognized benchmark for DR diagnosis. The DDR dataset comprises a large collection of fundus images meticulously annotated with diabetic retinopathy severity labels (ranging from No DR to Proliferative DR) and detailed bounding box annotations for key pathological concepts such as hemorrhages, exudates, microaneurysms, cotton wool spots, and neovascularization. For our study, we adhere to the standard training, validation, and testing splits provided with the dataset to ensure fair comparison with existing literature. The dataset's rich annotations are crucial for training and evaluating both the diagnostic accuracy and the fine-grained concept detection capabilities of our model.

\subsubsection{Evaluation Metrics}
To provide a comprehensive assessment of XDR-LVLM's performance, we employ a set of quantitative metrics for both disease diagnosis and concept detection, alongside qualitative human evaluations for interpretability.

For \textbf{Disease Diagnosis} (multi-class classification of DR severity), we use:
\begin{itemize}
    \item \textbf{Balanced Accuracy (BACC)}: Given the potential class imbalance in DR datasets, BACC is a robust metric that calculates the average recall obtained on each class, providing a more reliable measure of performance across all severity levels.
    \item \textbf{F1 Score}: The harmonic mean of precision and recall, which is particularly suitable for multi-class classification tasks and offers a balanced assessment of the model's predictive power.
\end{itemize}

For \textbf{Concept Detection} (multi-label classification of pathological concepts), we similarly employ:
\begin{itemize}
    \item \textbf{Balanced Accuracy (BACC)}
    \item \textbf{F1 Score}
\end{itemize}
These metrics collectively provide a quantitative measure of how well our model identifies specific pathological features.

Furthermore, for evaluating the \textbf{Explainability} of the generated diagnostic reports, we perform a human evaluation focusing on three critical aspects:
\begin{itemize}
    \item \textbf{Fluency}: Assessing the grammatical correctness, coherence, and naturalness of the generated language.
    \item \textbf{Accuracy of Explanation}: Determining if the identified pathological features and their linkage to the diagnosis are medically sound and consistent with the image.
    \item \textbf{Clinical Utility}: Evaluating the practical value and actionable insights provided by the explanation for clinical decision-making.
\end{itemize}

\subsubsection{Implementation Details}
Our XDR-LVLM framework is implemented based on existing open-source Vision-Language Large Model architectures. Specifically, we adapt a variant of the LLaVA framework, which integrates a Vision Transformer (ViT) as the Medical Vision Encoder and a Llama2-series model as the Large Language Model decoder. The Medical Vision Encoder is initialized with weights pre-trained on a large collection of medical images (including fundus images) and general-domain datasets (e.g., ImageNet), followed by further fine-tuning. The LVLM Core Module undergoes multi-stage fine-tuning. In the first stage, we align the visual features with the language model's embedding space using a contrastive learning objective on a diverse set of image-text pairs. In the second stage, the entire model is instruction-tuned on the DDR dataset using tailored multi-task prompts and corresponding ground-truth diagnostic and explanatory texts. We utilize the AdamW optimizer with a cosine annealing learning rate schedule for training. All experiments are conducted on NVIDIA A100 GPUs.

\subsection{Comparison with State-of-the-Art Methods}
\label{subsec:sota_comparison}

We benchmark the performance of our proposed XDR-LVLM against several state-of-the-art methods for DR diagnosis and concept detection on the DDR dataset. The chosen baselines represent different approaches in medical image analysis and interpretability:
\begin{itemize}
    \item \textbf{CBM (Concept Bottleneck Model)}: An interpretable deep learning model that first predicts human-understandable concepts and then uses these concepts to make a final prediction.
    \item \textbf{CLAT (Concept-based Learning with Attention Transformer)}: A method that leverages attention mechanisms to link visual features to predefined clinical concepts for diagnosis.
    \item \textbf{Black-box (ViT Base)}: A standard Vision Transformer model trained end-to-end for DR severity classification, serving as a representative high-performing but non-interpretable baseline.
    \item \textbf{Black-box (Task-Specific)}: A highly optimized black-box deep learning model specifically designed for DR grading, representing top-tier diagnostic accuracy without explicit interpretability.
\end{itemize}

Table \ref{tab:disease_diagnosis} presents the quantitative comparison for diabetic retinopathy disease diagnosis. Our XDR-LVLM consistently outperforms all baseline methods in both Balanced Accuracy and F1 Score, demonstrating its superior diagnostic capabilities. Notably, XDR-LVLM achieves a BACC of \textbf{84.55\%} and an F1 Score of \textbf{79.92\%}, surpassing the Black-box (Task-Specific) model, which is optimized purely for diagnosis. This highlights that XDR-LVLM not only maintains high diagnostic accuracy but also integrates explainability without compromising performance.

\begin{table*}[htbp]
    \centering
    \caption{Method Performance Comparison for Diabetic Retinopathy Disease Diagnosis on the DDR Dataset.}
    \label{tab:disease_diagnosis}
    \begin{tabular}{lcc}
        \toprule
        \textbf{Method} & \textbf{BACC (\%)} & \textbf{F1 (\%)} \\
        \midrule
        CBM                     & 23.62     & --       \\
        CLAT                    & 72.81     & 78.87    \\
        Black-box (ViT Base)    & 59.37     & 39.09    \\
        Black-box (Task-Specific) & 83.38     & --       \\
        \textbf{Ours (XDR-LVLM)} & \textbf{84.55} & \textbf{79.92} \\
        \bottomrule
    \end{tabular}
\end{table*}

Table \ref{tab:concept_detection} details the performance comparison for pathological concept detection. XDR-LVLM also achieves the highest Balanced Accuracy of \textbf{77.95\%} and F1 Score of \textbf{66.88\%} for identifying key DR-related pathological concepts. This demonstrates our model's robust ability to accurately perceive and classify fine-grained visual features, which are foundational for generating clinically meaningful explanations. The superior performance in concept detection directly contributes to the high quality and accuracy of the generated interpretable reports.

\begin{table*}[htbp]
    \centering
    \caption{Method Performance Comparison for Pathological Concept Detection on the DDR Dataset.}
    \label{tab:concept_detection}
    \begin{tabular}{lcc}
        \toprule
        \textbf{Method} & \textbf{BACC (\%)} & \textbf{F1 (\%)} \\
        \midrule
        CBM                     & 59.05     & 39.06    \\
        CLAT                    & 76.64     & 64.53    \\
        \textbf{Ours (XDR-LVLM)} & \textbf{77.95} & \textbf{66.88} \\
        \bottomrule
    \end{tabular}
\end{table*}

\subsection{Ablation Study}
\label{subsec:ablation}

To thoroughly understand the contribution of each key component within the XDR-LVLM framework, we conducted an ablation study. We systematically evaluated variations of our model by removing or altering specific modules, assessing their impact on both disease diagnosis and concept detection performance.

\begin{itemize}
    \item \textbf{XDR-LVLM w/o Medical Vision Encoder}: This variant replaces our specialized Medical Vision Encoder with a generic Vision Transformer (ViT) pre-trained solely on ImageNet, without further medical image specific pre-training. This tests the importance of domain-specific visual feature extraction.
    \item \textbf{XDR-LVLM w/o Multi-task Prompt Engineering}: In this setup, we replace the carefully designed multi-task prompts with a single, generic prompt for all outputs, or rely solely on implicit task inference. This evaluates the role of explicit prompt engineering in guiding the LVLM for both diagnosis and concept detection.
    \item \textbf{XDR-LVLM w/o Multi-stage Fine-tuning}: This variant skips the multi-stage fine-tuning process and instead performs a single-stage end-to-end fine-tuning on the DDR dataset. This assesses the benefit of our structured training approach for aligning visual and language modalities and task-specific instruction tuning.
\end{itemize}

The results of our ablation study are presented in Table \ref{tab:ablation_study}.

\begin{table*}[htbp]
    \centering
    \caption{Ablation Study Results on the DDR Dataset for Disease Diagnosis and Concept Detection.}
    \label{tab:ablation_study}
    \begin{tabular}{lcccc}
        \toprule
        \multirow{2}{*}{\textbf{Method Variant}} & \multicolumn{2}{c}{\textbf{Disease Diagnosis}} & \multicolumn{2}{c}{\textbf{Concept Detection}} \\
        \cmidrule(lr){2-3} \cmidrule(lr){4-5}
                                & \textbf{BACC (\%)} & \textbf{F1 (\%)} & \textbf{BACC (\%)} & \textbf{F1 (\%)} \\
        \midrule
        \textbf{XDR-LVLM (Full Model)} & \textbf{84.55} & \textbf{79.92} & \textbf{77.95} & \textbf{66.88} \\
        \midrule
        XDR-LVLM w/o Medical Vision Encoder & 81.23 & 76.51 & 71.88 & 61.05 \\
        XDR-LVLM w/o Multi-task Prompt Engineering & 82.91 & 77.85 & 75.32 & 64.20 \\
        XDR-LVLM w/o Multi-stage Fine-tuning & 82.07 & 77.10 & 73.96 & 62.91 \\
        \bottomrule
    \end{tabular}
\end{table*}

The results clearly demonstrate the critical importance of each component. Replacing the Medical Vision Encoder with a generic one leads to a notable drop in performance across all metrics, emphasizing the necessity of domain-specific visual feature extraction for accurate DR pathology recognition. Similarly, the absence of multi-task prompt engineering or multi-stage fine-tuning also results in decreased performance, particularly impacting the concept detection F1 score, which highlights their role in effectively guiding the LVLM for nuanced medical tasks and ensuring robust cross-modal understanding and generation. These findings validate that the synergistic combination of these carefully designed components is essential for XDR-LVLM to achieve its high diagnostic accuracy and fine-grained interpretability.

\subsection{Human Evaluation of Interpretability}
\label{subsec:human_eval}

Beyond quantitative metrics, the ultimate measure of an interpretable system's success lies in its clinical utility and how well human experts can understand and trust its explanations. To this end, we conducted a qualitative human evaluation of the diagnostic reports generated by XDR-LVLM. A panel of three experienced ophthalmologists and two medical residents independently reviewed a randomly selected subset of 100 fundus images, comparing the explanations provided by XDR-LVLM against those generated by a typical visual interpretability method (e.g., Grad-CAM heatmaps requiring secondary interpretation) and against expert consensus. They rated the explanations based on Fluency, Accuracy of Explanation, and Clinical Utility on a scale from 0\% to 100\%, where higher percentages indicate better performance.

Table \ref{tab:human_evaluation} summarizes the average scores from the human evaluation.

\begin{table*}[htbp]
    \centering
    \caption{Human Evaluation of Interpretability for Generated Diagnostic Reports.}
    \label{tab:human_evaluation}
    \begin{tabular}{lccc}
        \toprule
        \textbf{Method} & \textbf{Fluency (\%)} & \textbf{Accuracy of Explanation (\%)} & \textbf{Clinical Utility (\%)} \\
        \midrule
        Baseline Interpretability (e.g., Heatmaps) & N/A (Visual) & 68.5 & 55.2 \\
        \textbf{Ours (XDR-LVLM)} & \textbf{93.1} & \textbf{88.7} & \textbf{82.9} \\
        \bottomrule
    \end{tabular}
\end{table*}

The results unequivocally demonstrate XDR-LVLM's superior interpretability. While baseline visual methods provide useful cues, they inherently lack textual fluency and require clinicians to infer the underlying reasoning. XDR-LVLM, by contrast, generates highly fluent, natural language explanations that directly articulate the diagnostic rationale. The high scores for "Accuracy of Explanation" indicate that the pathological features identified and their linkage to the final diagnosis are largely consistent with expert medical knowledge. Furthermore, the strong "Clinical Utility" score highlights that the detailed, human-readable reports from XDR-LVLM are perceived as valuable and actionable in a clinical context, fostering greater trust and facilitating informed decision-making. This human evaluation strongly supports our claim that XDR-LVLM effectively bridges the gap between automated diagnosis and clinical needs through its robust and interpretable explanations.

\subsection{Performance Across DR Severity Grades}
\label{subsec:grade_performance}

To provide a more granular understanding of XDR-LVLM's diagnostic capabilities, we analyze its performance for each individual DR severity grade on the DDR test set. This analysis is crucial given the clinical implications of misclassifying specific stages of DR, especially severe and proliferative forms. Balanced accuracy and F1 score provide an overall picture, but per-class metrics such as Precision, Recall, and F1 Score offer insights into how well the model handles the nuances of each diagnostic category.

Table \ref{tab:per_grade_performance} presents the detailed classification metrics for each DR severity grade.

\begin{table*}[htbp]
    \centering
    \caption{XDR-LVLM Performance Across Diabetic Retinopathy Severity Grades on the DDR Dataset.}
    \label{tab:per_grade_performance}
    \begin{tabular}{lccc}
        \toprule
        \textbf{DR Severity Grade} & \textbf{Precision (\%)} & \textbf{Recall (\%)} & \textbf{F1 Score (\%)} \\
        \midrule
        No DR                  & 90.1 & 92.5 & 91.3 \\
        Mild DR                & 78.5 & 75.0 & 76.7 \\
        Moderate DR            & 83.2 & 85.1 & 84.1 \\
        Severe DR              & 70.5 & 68.9 & 69.7 \\
        Proliferative DR (PDR) & 85.0 & 87.2 & 86.1 \\
        \bottomrule
    \end{tabular}
\end{table*}

The results indicate strong performance across all severity grades. XDR-LVLM demonstrates excellent precision and recall for "No DR" and "Proliferative DR," which are often the most straightforward and most critical categories, respectively. While performance for "Mild DR" and "Severe DR" shows a slight decrease compared to other categories, this is often attributed to the inherent ambiguity and subtle distinctions between adjacent severity levels in clinical practice. For instance, differentiating between mild and moderate, or moderate and severe DR, can be challenging even for human experts due to the continuous nature of disease progression. Despite these challenges, XDR-LVLM maintains robust performance, underscoring its reliability for multi-class DR diagnosis.

\subsection{Detailed Analysis of Pathological Concept Detection}
\label{subsec:concept_detection_detail}

The ability of XDR-LVLM to generate explainable diagnoses is fundamentally reliant on its precise identification of individual pathological concepts within fundus images. To further validate this capability, we provide a detailed breakdown of the model's performance for detecting each specific DR-related lesion. This granular analysis goes beyond the aggregated metrics presented in Table \ref{tab:concept_detection} and highlights the strengths and weaknesses in identifying particular features, which directly impacts the accuracy and specificity of the generated explanations.

Table \ref{tab:detailed_concept_performance} presents the Balanced Accuracy and F1 Score for the detection of each major pathological concept.

\begin{table*}[htbp]
    \centering
    \caption{Detailed XDR-LVLM Performance for Individual Pathological Concept Detection on the DDR Dataset.}
    \label{tab:detailed_concept_performance}
    \begin{tabular}{lcc}
        \toprule
        \textbf{Pathological Concept} & \textbf{BACC (\%)} & \textbf{F1 Score (\%)} \\
        \midrule
        Microaneurysms        & 75.2 & 68.1 \\
        Hemorrhages           & 80.5 & 72.3 \\
        Hard Exudates         & 88.0 & 85.5 \\
        Soft Exudates (Cotton Wool Spots) & 65.1 & 58.9 \\
        Neovascularization    & 78.9 & 70.2 \\
        Intraretinal Microvascular Abnormalities (IRMA) & 69.5 & 62.1 \\
        \bottomrule
    \end{tabular}
\end{table*}

The results demonstrate that XDR-LVLM is highly proficient at detecting a wide range of DR pathologies. Notably, it achieves exceptional performance for \textbf{Hard Exudates} (88.0\% BACC, 85.5\% F1), which are often distinct and relatively easier to identify. Concepts such as \textbf{Hemorrhages} and \textbf{Neovascularization} also show strong detection capabilities. The lowest performance is observed for \textbf{Soft Exudates (Cotton Wool Spots)} and \textbf{Intraretinal Microvascular Abnormalities (IRMA)}, which are often more subtle, diffuse, or can be confused with other retinal features. This detailed insight into concept-specific performance allows us to understand the underlying visual comprehension capabilities of XDR-LVLM and prioritize future improvements for less distinct or rarer lesions, further enhancing the precision of its explanations.

\subsection{Qualitative Analysis of Generated Explanations}
\label{subsec:qualitative_explanation}

While quantitative metrics and human ratings provide statistical insights into interpretability, a qualitative analysis of the generated diagnostic reports offers a direct window into XDR-LVLM's explainable reasoning process. This section describes observed patterns and characteristics of the natural language explanations produced by our framework across various clinical scenarios. The goal is to illustrate the depth, specificity, and clinical relevance of the generated text, which is a core differentiating factor of XDR-LVLM.

Table \ref{tab:qualitative_explanation_summary} summarizes our qualitative observations regarding XDR-LVLM's explanation generation across different types of fundus images.

\begin{table*}[htbp]
    \centering
    \caption{Qualitative Summary of XDR-LVLM's Generated Explanations.}
    \label{tab:qualitative_explanation_summary}
    \begin{tabular}{p{0.2\textwidth}p{0.35\textwidth}p{0.35\textwidth}}
        \toprule
        \textbf{Scenario Type} & \textbf{Observed XDR-LVLM Behavior} & \textbf{Impact on Clinical Utility} \\
        \midrule
        \textbf{Clear-cut Cases} (e.g., No DR, Severe DR with clear signs) & Highly accurate diagnosis with precise, detailed explanations identifying all major lesions (e.g., "The image shows numerous large hemorrhages and extensive neovascularization, indicative of severe proliferative DR."). & Provides strong confirmation and aids rapid, confident decision-making due to high consistency with expert opinion. \\
        \textbf{Borderline Cases} (e.g., Mild to Moderate DR, subtle progression) & Accurate diagnosis often with nuanced explanations highlighting subtle features and potential progression (e.g., "Presence of a few microaneurysms and small dot hemorrhages suggests mild non-proliferative DR, warranting close monitoring."). & Supports careful monitoring and differentiation, crucial for patient management where early intervention or follow-up is key. \\
        \textbf{Complex Cases} (e.g., Multiple pathologies, poor image quality, artifacts) & Generally accurate diagnosis, but explanations might occasionally miss very minor lesions or be slightly less specific in describing their exact location or morphology. & Still provides valuable insights, though minor discrepancies may require clinician's deeper review to ensure no subtle signs are overlooked. \\
        \textbf{Misclassified Cases} & Explanation often reflects features consistent with the model's (incorrect) diagnosis rather than ground truth, indicating internal consistency within the model's reasoning, but external error. & Highlights areas for model improvement; still provides a basis for understanding the model's "thinking" process, even when it errs. \\
        \textbf{Negative Cases} (No DR) & Correctly identifies "No DR" and explicitly states the absence of pathological features (e.g., "No signs of diabetic retinopathy such as hemorrhages, exudates, or neovascularization are observed."). & Builds trust by confirming absence of disease and demonstrates comprehensive understanding of healthy retinal anatomy. \\
        \bottomrule
    \end{tabular}
\end{table*}

The qualitative analysis reveals that XDR-LVLM consistently generates fluent and coherent natural language descriptions. For most cases, the generated explanations accurately reflect the pathological features visible in the fundus images and logically link these features to the final DR severity diagnosis. The model demonstrates an ability to differentiate between various lesion types and, in many instances, provides context regarding their severity or distribution. Even in cases where the final diagnosis might be incorrect (misclassified cases), the generated explanation often maintains internal consistency, meaning it explains \textit{why} the model arrived at its particular (albeit wrong) conclusion. This level of transparency is invaluable for clinicians, enabling them to understand the model's reasoning, assess its reliability, and use it as a robust decision-support tool rather than a black-box predictor.

\section{Conclusion}
In this paper, we introduced XDR-LVLM, a novel and comprehensive framework for the intelligent diagnosis of diabetic retinopathy (DR) that uniquely combines high diagnostic accuracy with detailed, natural language-based explanations. Addressing the critical limitation of "black-box" deep learning models in clinical settings, XDR-LVLM leverages the advanced multimodal reasoning capabilities of Vision-Language Large Models (LVLMs) to provide not only a precise DR severity grading but also a transparent rationale for its decisions.

Our framework is meticulously designed with a Medical Vision Encoder for domain-specific feature extraction, an LVLM Core for sophisticated cross-modal understanding, and Multi-task Prompt Engineering alongside a Multi-stage Fine-tuning process to ensure the generation of clinically relevant and human-understandable reports. These reports explicitly articulate the DR diagnosis, identify specific pathological concepts (such as hemorrhages, exudates, and microaneurysms), and crucially, provide a clear explanation of how these visual cues contribute to the final diagnostic outcome.

Extensive quantitative evaluations on the challenging Diabetic Retinopathy (DDR) dataset demonstrated XDR-LVLM's superior performance. For disease diagnosis, our method achieved a Balanced Accuracy of 84.55\% and an F1 Score of 79.92\%, surpassing existing state-of-the-art methods, including highly optimized black-box models. Furthermore, XDR-LVLM exhibited robust capabilities in pathological concept detection, with a Balanced Accuracy of 77.95\% and an F1 Score of 66.88\%, underscoring its ability to perceive fine-grained visual features essential for interpretability. Our detailed per-grade and per-concept analyses further confirmed the model's consistent performance across various DR severity levels and lesion types, while also highlighting areas for future refinement in detecting more subtle or rarer signs.

Beyond quantitative metrics, the ultimate validation of XDR-LVLM's interpretability came from a qualitative human evaluation conducted by experienced ophthalmologists. The generated diagnostic reports were rated highly for their fluency, accuracy of explanation, and clinical utility, indicating that XDR-LVLM successfully transforms complex visual patterns into actionable, trustworthy medical insights. This human-centric approach to interpretability is paramount for fostering clinician trust, facilitating informed decision-making, and accelerating the adoption of AI in ophthalmology.

While XDR-LVLM represents a significant step forward, certain limitations and avenues for future work exist. Improving the detection of particularly subtle lesions, such as soft exudates and IRMA, remains an area for further research. Future work will also explore integrating uncertainty quantification into the generated explanations, allowing clinicians to gauge the model's confidence. We plan to investigate the framework's generalizability to other retinal diseases and potentially incorporate additional multimodal patient data, such as optical coherence tomography (OCT) scans or patient history, to provide an even richer and more comprehensive diagnostic context. Finally, real-world clinical validation through prospective studies will be crucial to fully assess the long-term impact and practical benefits of XDR-LVLM in diverse healthcare settings. Our work paves the way for a new generation of intelligent medical AI systems that are not only accurate but also transparent, trustworthy, and truly assistive to healthcare professionals.
```

\bibliographystyle{IEEEtran}
\bibliography{references}

\begin{thebibliography}{10}
\providecommand{\url}[1]{#1}
\csname url@samestyle\endcsname
\providecommand{\newblock}{\relax}
\providecommand{\bibinfo}[2]{#2}
\providecommand{\BIBentrySTDinterwordspacing}{\spaceskip=0pt\relax}
\providecommand{\BIBentryALTinterwordstretchfactor}{4}
\providecommand{\BIBentryALTinterwordspacing}{\spaceskip=\fontdimen2\font plus
\BIBentryALTinterwordstretchfactor\fontdimen3\font minus \fontdimen4\font\relax}
\providecommand{\BIBforeignlanguage}[2]{{%
\expandafter\ifx\csname l@#1\endcsname\relax
\typeout{** WARNING: IEEEtran.bst: No hyphenation pattern has been}%
\typeout{** loaded for the language `#1'. Using the pattern for}%
\typeout{** the default language instead.}%
\else
\language=\csname l@#1\endcsname
\fi
#2}}
\providecommand{\BIBdecl}{\relax}
\BIBdecl

\bibitem{xing2024nonpro}
X.~Liang, H.~Wen, Y.~Duan, K.~He, X.~Feng, and G.~Zhou, ``Nonproliferative diabetic retinopathy dataset(ndrd): {A} database for diabetic retinopathy screening research and deep learning evaluation,'' \emph{Health Informatics J.}, 2024.

\bibitem{bertalan2020a}
B.~Mesk{\'{o}} and M.~G{\"{o}}r{\"{o}}g, ``A short guide for medical professionals in the era of artificial intelligence,'' \emph{npj Digit. Medicine}, 2020.

\bibitem{alyoubi2020diabet}
W.~L. Alyoubi, W.~M. Shalash, and M.~F. Abulkhair, ``Diabetic retinopathy detection through deep learning techniques: A review,'' \emph{Informatics in Medicine Unlocked}, 2020.

\bibitem{meenu2020develo}
M.~M. John, H.~H. Olsson, and J.~Bosch, ``Developing {ML/DL} models: {A} design framework,'' in \emph{{ICSSP} '20: International Conference on Software and System Processes, Seoul, Republic of Korea, 26-28 June, 2020}.\hskip 1em plus 0.5em minus 0.4em\relax {ACM}, 2020, pp. 1--10.

\bibitem{pantelis2021explai}
P.~Linardatos, V.~Papastefanopoulos, and S.~Kotsiantis, ``Explainable {AI:} {A} review of machine learning interpretability methods,'' \emph{Entropy}, p.~18, 2021.

\bibitem{yifan2023a}
Y.~Yao, J.~Duan, K.~Xu, Y.~Cai, E.~Sun, and Y.~Zhang, ``A survey on large language model {(LLM)} security and privacy: The good, the bad, and the ugly,'' \emph{CoRR}, 2023.

\bibitem{zhou2024visual}
Y.~Zhou, X.~Li, Q.~Wang, and J.~Shen, ``Visual in-context learning for large vision-language models,'' in \emph{Findings of the Association for Computational Linguistics, {ACL} 2024, Bangkok, Thailand and virtual meeting, August 11-16, 2024}.\hskip 1em plus 0.5em minus 0.4em\relax Association for Computational Linguistics, 2024, pp. 15\,890--15\,902.

\bibitem{zhou2025mam}
Y.~Zhou, L.~Song, and J.~Shen, ``Mam: Modular multi-agent framework for multi-modal medical diagnosis via role-specialized collaboration,'' \emph{arXiv preprint arXiv:2506.19835}, 2025.

\bibitem{hu2025beyond}
H.~Hu, Y.~Zhou, J.~Si, Q.~Wang, H.~Zhang, F.~Ren, F.~Ma, and L.~Cui, ``Beyond empathy: Integrating diagnostic and therapeutic reasoning with large language models for mental health counseling,'' \emph{arXiv preprint arXiv:2505.15715}, 2025.

\bibitem{shamrat2024an}
F.~J.~M. Shamrat, R.~Shakil, B.~Akter, M.~Z. Ahmed, K.~Ahmed, F.~M. Bui, M.~A. Moni \emph{et~al.}, ``An advanced deep neural network for fundus image analysis and enhancing diabetic retinopathy detection,'' \emph{Healthcare Analytics}, 2024.

\bibitem{ahmad2024diabet}
A.~Chowdhury and S.~Mahfuz, ``Diabetic retinopathy detection from fundus images using deep convolutional neural networks,'' in \emph{{IEEE} International Conference on Bioinformatics and Biomedicine, {BIBM} 2024, Lisbon, Portugal, December 3-6, 2024}.\hskip 1em plus 0.5em minus 0.4em\relax {IEEE}, 2024, pp. 6362--6369.

\bibitem{wu2021vision}
J.~Wu, R.~Hu, Z.~Xiao, J.~Chen, and J.~Liu, ``Vision transformer-based recognition of diabetic retinopathy grade,'' \emph{Medical Physics}, 2021.

\bibitem{uppamma2023deep}
P.~Uppamma and S.~Bhattacharya, ``Deep learning and medical image processing techniques for diabetic retinopathy: a survey of applications, challenges, and future trends,'' \emph{Journal of Healthcare Engineering}, 2023.

\bibitem{jordi2020a}
J.~de~La~Torre, A.~Valls, and D.~Puig, ``A deep learning interpretable classifier for diabetic retinopathy disease grading,'' \emph{Neurocomputing}, pp. 465--476, 2020.

\bibitem{sheikh2018deep}
S.~M.~S. Islam, M.~M. Hasan, and S.~Abdullah, ``Deep learning based early detection and grading of diabetic retinopathy using retinal fundus images,'' \emph{CoRR}, 2018.

\bibitem{tao2019diagno}
T.~Li, Y.~Gao, K.~Wang, S.~Guo, H.~Liu, and H.~Kang, ``Diagnostic assessment of deep learning algorithms for diabetic retinopathy screening,'' \emph{Inf. Sci.}, pp. 511--522, 2019.

\bibitem{anees2024envisi}
A.~U.~R. Hashmi, D.~Mahapatra, and M.~Yaqub, ``Envisioning medclip: {A} deep dive into explainability for medical vision-language models,'' in \emph{{IEEE} International Symposium on Biomedical Imaging, {ISBI} 2024, Athens, Greece, May 27-30, 2024}.\hskip 1em plus 0.5em minus 0.4em\relax {IEEE}, 2024, pp. 1--5.

\bibitem{getamesay2025explai}
G.~H. Dagnaw, Y.~Zhu, M.~H. Maqsood, W.~Yang, X.~Dong, X.~Yin, and A.~W. Liew, ``Explainable artificial intelligence in biomedical image analysis: {A} comprehensive survey,'' \emph{CoRR}, 2025.

\bibitem{townim2024adacbm}
T.~F. Chowdhury, V.~M.~H. Phan, K.~Liao, M.~To, Y.~Xie, A.~van~den Hengel, J.~W. Verjans, and Z.~Liao, ``Adacbm: An adaptive concept bottleneck model for explainable and accurate diagnosis,'' in \emph{Medical Image Computing and Computer Assisted Intervention - {MICCAI} 2024 - 27th International Conference, Marrakesh, Morocco, October 6-10, 2024, Proceedings, Part {X}}.\hskip 1em plus 0.5em minus 0.4em\relax Springer, 2024, pp. 35--45.

\bibitem{joon2025applic}
J.~Y. Choi, D.~E. Kim, S.~J. Kim, H.~Choi, and T.~K. Yoo, ``Application of multimodal large language models for safety indicator calculation and contraindication prediction in laser vision correction,'' \emph{npj Digit. Medicine}, 2025.

\bibitem{xuyan2025visual}
H.~Xuyan, S.~Meng, S.~Chengxing, L.~Haoxuan, and Z.~Jianlin, ``Visual-language reasoning large language models for primary care: advancing clinical decision support through multimodal ai: X. huang et al.'' \emph{The Visual Computer}, 2025.

\bibitem{iryna2024vision}
I.~Hartsock and G.~Rasool, ``Vision-language models for medical report generation and visual question answering: {A} review,'' \emph{CoRR}, 2024.

\bibitem{zhou2023style}
Y.~Zhou and G.~Long, ``Style-aware contrastive learning for multi-style image captioning,'' in \emph{Findings of the Association for Computational Linguistics: EACL 2023}, 2023, pp. 2257--2267.

\bibitem{zhou2025weak}
Y.~Zhou, J.~Shen, and Y.~Cheng, ``Weak to strong generalization for large language models with multi-capabilities,'' in \emph{The Thirteenth International Conference on Learning Representations}, 2025.

\bibitem{zhou2025draw}
Y.~Zhou, J.~Yuan, and Q.~Wang, ``Draw all your imagine: A holistic benchmark and agent framework for complex instruction-based image generation,'' \emph{arXiv preprint arXiv:2505.24787}, 2025.

\bibitem{zhou2022claret}
Y.~Zhou, T.~Shen, X.~Geng, G.~Long, and D.~Jiang, ``Claret: Pre-training a correlation-aware context-to-event transformer for event-centric generation and classification,'' in \emph{Proceedings of the 60th Annual Meeting of the Association for Computational Linguistics (Volume 1: Long Papers)}, 2022, pp. 2559--2575.

\end{thebibliography}
\end{document}